\documentclass{article}

\usepackage{amsmath,graphicx}
\usepackage{import}

\usepackage{graphicx}
\usepackage{xspace}
\usepackage{pdfpages}
\usepackage{amsmath,amssymb,amsfonts}
\usepackage{bbm}
\usepackage{bm}
\usepackage{diagbox}
\usepackage[pagebackref,breaklinks,colorlinks]{hyperref}
\hypersetup{
    colorlinks=true, 
    linkcolor=blue,  
    filecolor=magenta, 
    urlcolor=blue,    
    citecolor=blue,       
}
\usepackage{enumitem}
\usepackage{color,soul}
\usepackage{xcolor}  
\usepackage{pifont}
\usepackage{tikz}
\usetikzlibrary{positioning,fit,shapes.geometric,arrows.meta}
\usetikzlibrary{calc}
\usepackage{pgfplots}
\definecolor{lineBlue}{RGB}{57,106,177}
\definecolor{lineOrange}{RGB}{218,124,48}
\definecolor{lineGreen}{RGB}{62,150,81}
\definecolor{lineRed}{RGB}{204,37,41}
\definecolor{lineGray}{RGB}{83,81,84}
\definecolor{linePurple}{RGB}{107,76,154}
\definecolor{lineMaroon}{RGB}{146,36,40}
\definecolor{barBlue}{RGB}{114,147,203}
\definecolor{barOrange}{RGB}{225,151,76}
\definecolor{barGreen}{RGB}{132,186,91}
\definecolor{barRed}{RGB}{211,94,96}
\definecolor{barGray}{RGB}{128,133,133}
\definecolor{barPurple}{RGB}{144,103,167}
\definecolor{barMaroon}{RGB}{171,104,81}
\pgfplotsset{compat=1.18}
\usetikzlibrary{3d,decorations.text,shapes.arrows,positioning,fit,backgrounds}
\usetikzlibrary{intersections, pgfplots.fillbetween}
\usepackage{multirow}
\usetikzlibrary{arrows.meta}
\usepackage{algorithm}
\usepackage{algorithmic}

\subimport{./}{macros}

\usepackage[numbers]{natbib}
\usepackage[final]{neurips_2025}

\usepackage[utf8]{inputenc} 
\usepackage[T1]{fontenc}    
\usepackage{hyperref}       
\usepackage{url}            
\usepackage{booktabs}       
\usepackage{amsfonts}       
\usepackage{nicefrac}       
\usepackage{microtype}      
\usepackage{xcolor}         
\usepackage{import}
\usepackage{graphicx}

\usepackage{caption}

\title{FastJAM: a Fast Joint Alignment Model for Images}

\author{
\hypersetup{hidelinks}
  Omri Hirsch\thanks{Equal Contribution}\hspace{0.7em}
  Ron Shapira Weber\footnotemark[1]\hspace{0.7em}
  Shira Ifergane\hspace{0.7em}
  Oren Freifeld \\
  The Faculty of Computer and Information Science, Ben Gurion University of the Negev (BGU), Israel \\
  The Data Science Research Center, BGU \\
  The School of Brain Sciences and Cognition, BGU \\  \\
  \texttt{\{omrihir,ronsha,shiraif\}@post.bgu.ac.il} \\
  \texttt{orenfr@bgu.ac.il}
}

\begin{document}

\maketitle

\begin{abstract}
Joint Alignment (JA) of images aims to align a collection of images into a unified coordinate frame, such that semantically-similar features appear at corresponding spatial locations. Most existing approaches often require long training times, large-capacity models,
and extensive hyperparameter tuning. We introduce FastJAM, a rapid, graph-based method that drastically reduces the computational complexity of joint alignment tasks. FastJAM leverages pairwise matches computed by an off-the-shelf image matcher, together with a rapid nonparametric clustering,  to construct a graph representing intra- and inter-image keypoint relations. 
A graph neural network propagates and aggregates these correspondences, efficiently predicting per-image homography parameters via image-level pooling. Utilizing an inverse-compositional loss, that eliminates the need
for a regularization term over the predicted transformations (and thus
also obviates the hyperparameter tuning associated with such terms), FastJAM performs image JA quickly and effectively. Experimental results on several benchmarks demonstrate that FastJAM achieves results better than existing modern JA methods in terms of alignment quality, while reducing computation time from hours or minutes to mere seconds. Our code is available at our project webpage, \url{https://bgu-cs-vil.github.io/FastJAM/}.
\end{abstract}

\section{Introduction}\label{Sec:Intro}
\begin{figure}[t]
    \centering
    \includegraphics[trim=90mm 15mm 75mm 15mm, clip, width=0.9\linewidth]{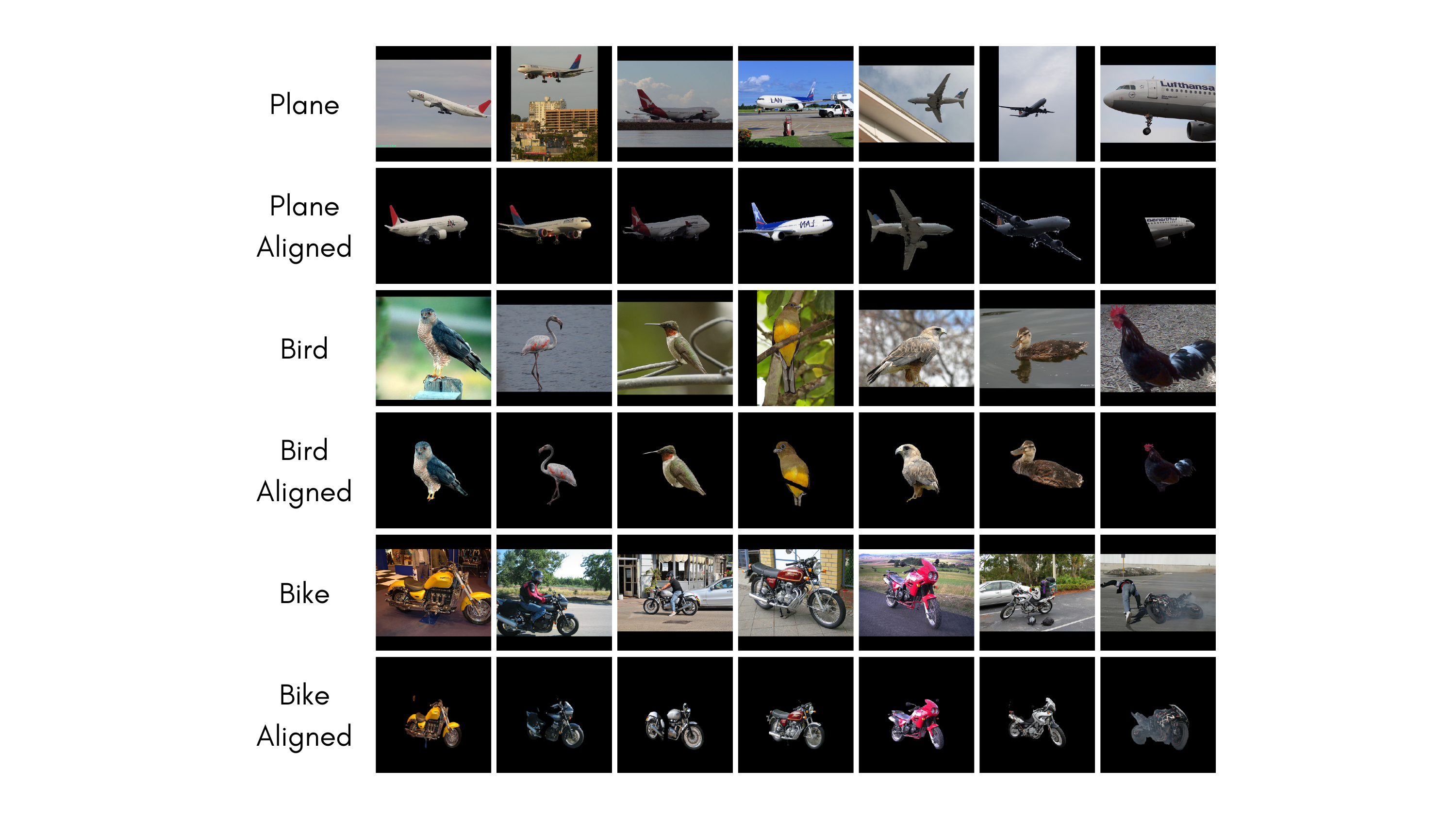}
    \caption{\textbf{Joint alignment with FastJAM.} Given a set of images of the same object, or of different objects from the same category (\eg, motorbikes), our method aligns all images in seconds, compared to other methods (minutes~\cite{Barel:ECCV:2024:spacejam} or hours~\cite{Ofri:CVPR:2023:neuracongealingl,Gupta:ICCV:2023:ASIC}).}
    \label{Fig:Alig}
\end{figure}

Joint Alignment (JA) is the task of aligning a collection of images by estimating per-image spatial transformations such that, when applied, all images become geometrically consistent in a shared coordinate frame according to certain semantic or geometric criteria (see, \eg,~\autoref{Fig:Alig}). Unlike pairwise alignment, which aligns each image pair independently and often leads to error accumulation (\ie, ``drifting''),
JA enforces a global agreement across the entire set,  making JA particularly valuable for discovering shared structures between images or building a class atlas.
However, achieving JA is inherently challenging: without supervision or a reference image, optimization methods frequently collapse into trivial or inconsistent solutions. Moreover, existing approaches typically require extensive computational resources, taking more than an hour~\cite{Ofri:CVPR:2023:neuracongealingl,Gupta:ICCV:2023:ASIC}  to jointly align as few as 30 images. Recently, we proposed a method called \textbf{SpaceJAM}~\cite{Barel:ECCV:2024:spacejam} that, partially by virtue of a new \emph{inverse-compositional loss over dense feature maps}, significantly mitigated these computational issues, thereby solved the task in only a few minutes. Additionally, SpaceJAM set new state-of-the-art  
quantitative results. 
A natural question arises, however: is it possible to do even better
in terms of both speed and performance? Fortunately, the answer is positive, as we show in the present paper.

Here we introduce an even more computationally-efficient method that solves the JA problem in under 50 seconds, dramatically outperforming prior approaches in terms of speed while maintaining, and in fact typically improving, alignment quality.

Traditionally, JA methods relied on classical approaches such as congealing~\cite{Miller:CVPR:2000:learning,Learned:PAMI:2006:align}, which iteratively align each image towards the remaining set, or centroid-based methods that utilize a reference image or a latent template~\cite{Barel:ECCV:2024:spacejam}.
Classical techniques employed feature-based methods (\eg, SIFT~\cite{Lowe:ICCV:1999:SIFT}), to establish keypoint (KP) correspondences between images. The rise of deep learning, particularly through Vision Transformers (ViTs)~\cite{Dosovitskiy:2020:ViT} and semantic feature extraction methods like DINO~\cite{Caron:ICCV:2021:VITDINO}, has significantly advanced JA by providing richer representations that alleviate some challenges faced by traditional methods. However, even with ViT features, many difficulties persist, leading recent approaches to depend heavily on high-capacity, computationally-expensive models paired with extensive regularization~\cite{Ofri:CVPR:2023:neuracongealingl, Gupta:ICCV:2023:ASIC}. This reliance not only increases computational demands but also introduces complexity through the requirement of extensive hyperparameter (HP) tuning, ultimately resulting in methods that are slow and often brittle (as the HP tuning is usually dataset-specific). 

Both congealing and atlas-based approaches typically rely on objective functions that fall into two main categories: geometric losses and semantic (feature-based) losses. In the geometric cases, one directly minimizes spatial discrepancies between corresponding KPs across images, leveraging explicit correspondence information. In contrast, semantic losses operate over dense feature representations and provide a smoother, globally-differentiable alignment without requiring explicit KPs.
However, such dense (and typically high-dimensional) representations, often derived from high-capacity models like DINO~\cite{Caron:ICCV:2021:VITDINO}, are computationally expensive when used within the JA optimization (even if the features themselves are kept frozen). Therefore, in this work we adopt the geometric loss paradigm, offering a 
sparse, lightweight, and scalable formulation that achieves typically higher alignment accuracy while significantly reducing the computational overhead.
 
\begin{figure}[t]
    \centering
    \includegraphics[trim=0mm 10mm 0mm 0mm, clip, width=0.9\linewidth]{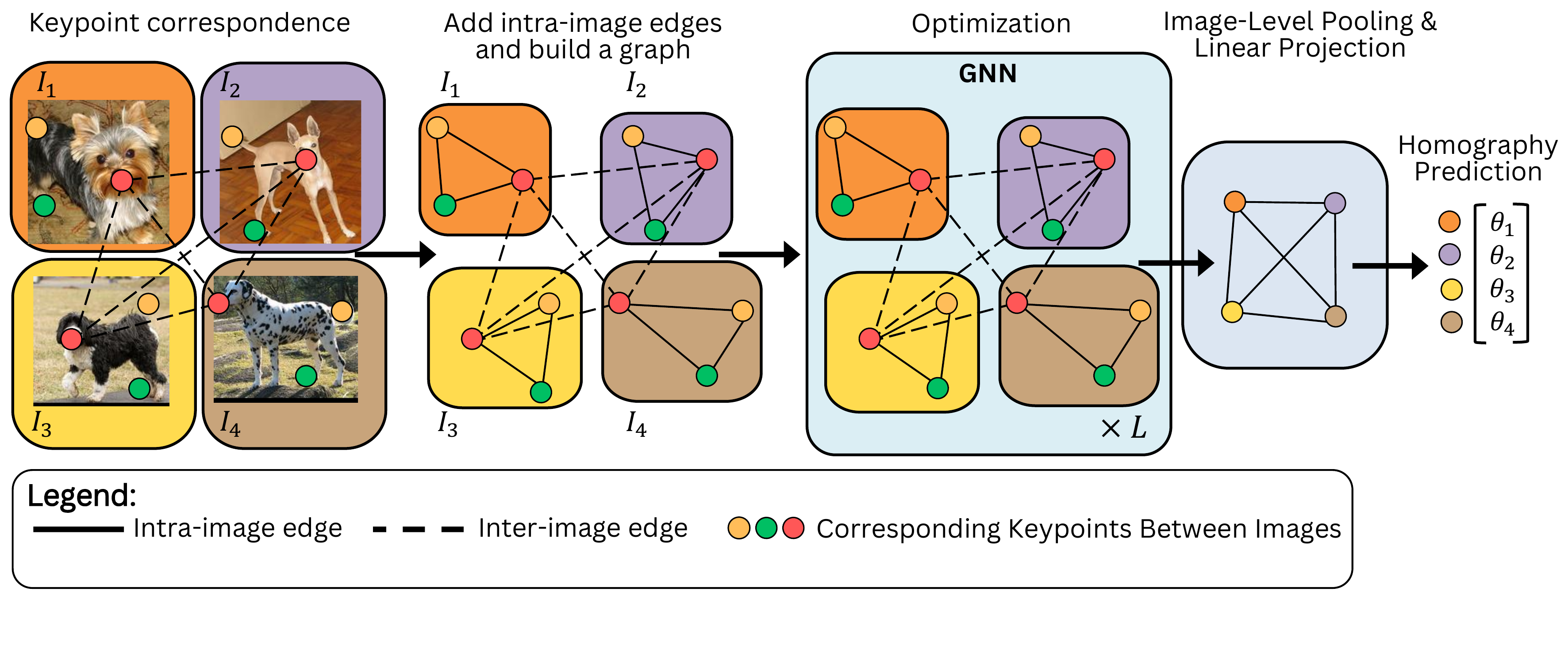}
    \caption{\textbf{Overview of the FastJAM architecture.} Given a set of images, we extract sparse keypoints (KPs)  and pairwise correspondences using an off-the-shelf matcher (left; only red-dot matches are shown for clarity). A graph is built by linking KPs within each image (intra-image edges) and across matched pairs (inter-image edges). A GNN with $L$ layers propagates alignment information through this graph (center). Image-level features are then obtained via mean pooling and used to predict per-image homography parameters $(\btheta_i)$ for joint alignment.}
    \label{fig:fastjam:arch}
\end{figure}

 Concretely, we introduce \textbf{FastJAM}, a graph-based JA framework that achieves fast and scalable alignment. Unlike prior methods that rely on dense feature maps and/or computationally-intensive optimization, FastJAM constructs a KP graph from pairwise correspondences using an off-the-shelf matcher, where nodes represent KPs and edges encode intra- and inter-image relationships. A Graph Neural Network (GNN) propagates alignment cues across the graph, and a readout layer produces image-level embeddings used to predict the homography parameters (as shown in~\autoref{fig:fastjam:arch}). Combined with a robust \emph{inverse-compositional geometric loss}, FastJAM aligns an entire image set in under 50 seconds. Experiments on SPair-71k and CUB-200 show that FastJAM matches or exceeds the accuracy of contemporary JA methods while being significantly more efficient and 
 orders of magnitude faster (see~\autoref{tab:ja:comparison} for a comparison).
Our key contributions are as follows. 
\begin{enumerate}
    \item We introduce FastJAM, a novel GNN-based framework for JA that significantly accelerates the alignment process (compared with existing methods) from hours/minutes to seconds.
    \item FastJAM graph structure allows for the information from the entire image collection to propagate between images during optimization, unlike previous ``image-by-image'' approaches (including SpaceJAM), leading to improvements in JA quality. 
    \item The first Inverse-Compositional JA loss that is based on a sparse KP representation. 
\end{enumerate}

\section{Related Work}\label{Sec:Related}

\begin{table}[t]
\centering
\caption{Comparison with recent JA methods on three SPair-71k categories~\cite{Min:2019:spair}. Runtime is reported as average $\pm$ standard deviation in \texttt{hh:mm:ss} format.}
\small
\label{tab:ja:comparison}
\begin{tabular}{lccccc|c}
\toprule
\textbf{Method} & \# Params & \# Losses & \#HP & Atlas-free & \# Epochs & Runtime \\
\midrule
Neural Congealing~\cite{Ofri:CVPR:2023:neuracongealingl} & 28.7M & 8 & 8 & \xmark & 8000 & 01:18:30 $\pm$ 00:06:18 \\
ASIC~\cite{Gupta:ICCV:2023:ASIC} & 7.9M & 4 & 5 & \xmark & 20000 & 01:06:38 $\pm$ 00:00:38 \\
SpaceJAM~\cite{Barel:ECCV:2024:spacejam} & 0.016M & 1 & 0 & \cmark & 700 & 00:06:00 $\pm$ 00:00:12 \\
FastJAM (Ours) & 0.13M & 1 & 0 & \cmark & 600 & \textbf{00:00:49} $\pm$ 00:00:04 \\
\bottomrule
\end{tabular}
\end{table}


\paragraph{Pairwise image alignment.}
Learning-based correspondence methods have substantially improved the accuracy and robustness of image matching. Sparse approaches like SuperPoint~\cite{DeTone:CVPR:2018:superpoint} and SuperGlue~\cite{Sarlin:CVPR:2020:superglue} detect KPs and compute context-aware matches using attention and GNNs, but might struggle in low-texture or repetitive regions due to their reliance on sparse detections. To overcome these limitations, dense methods such as LoFTR~\cite{Sun:CVPR:2021:loftr} compute pixel-wise matches using transformer-based architectures without requiring explicit KPs. Recent advances continue to close the gap between the sparse and dense paradigms. RoMa~\cite{Edstedt:CVPR:2024:RoMa} combines DINOv2 features with hierarchical transformers for robust wide-baseline matching. DIFT~\cite{Tang:NIPS:2023:DIFT} introduces efficient descriptor interpolation from diffusion models. Additional types of relevant dense features appeared in the works of  Mariotti~\etal~\cite{Mariotti:CVPR:2024:improving}, who propose spherical viewpoint maps that encode rich geometry, and Xu~\etal~\cite{Zhang:CVPR:2024:telling}, who incorporate directional priors to disambiguate symmetric cases.
However, common issues with dense matching approaches are that they 
can be computationally demanding and that they require post-processing because the geometry often breaks. 

\paragraph{JA by feature matching.}
Classical JA methods utilize geometric transformations and handcrafted features. The idea of  congealing~\cite{Miller:CVPR:2000:learning,Learned:PAMI:2006:align} 
is to iteratively align images by minimizing a global cost, typically entropy or least-squares of pixel values~\cite{Cox:CVPR:2008:LS,Cox:ICCV:2009:LS,Learned:PAMI:2006:align} or descriptors like SIFT~\cite{Huang:CVPR:2007:unsupervised,Lin:CVPR:2012:aligning,Shokrollahi:CVPR:2015:aligngraph}. Other approaches simultaneously cluster and align images to their class means~\cite{Liu:ICCV:2009:simultaneous,Mattar:2012:unsupervised} or use template matching~\cite{Jain:TPAMI:1996:objectTemplate,Gavrila:ICPR:1998:multi,Felzenszwalb:CVPR:2007:hierarchical}. 
In any case, such methods are limited by the quality of the extracted features.
Another classical approach models image sets as low-rank linear subspaces~\cite{Kemelmacher:CVPR:2012:collectionflow,He:IVC:2014:GRASTA,Zhang:TPAMI:2019:robust, Peng:TPAMIL2012:RASL}.

Deep learning has substantially advanced image JA. Huang~\etal~\cite{Huang:NIPS:2012:learning} adapted congealing to CNN features, while Spatial Transformer Networks (STNs)~\cite{Jaderberg:NIPS:2015:STN} introduced a differentiable module for predicting spatial transformations, enabling end-to-end alignment learning. STNs have since been widely adopted in JA tasks, including congealing~\cite{Annunziata:ICCV:2019:jointly}, atlas construction~\cite{Lin:CPVR:2017:inverse,Dalca:NIPS:2019:learning,Shapira:NIPS:2019:DTAN,Sinclair:MIA:2022:ISTN,Shapira:ICML:2023:RFDTAN}, joint clustering~\cite{Monnier:NIPS:2020:DTIC,Loiseau:ICV:2021:representing},  moving-camera background modeling~\cite{Chelly:CVPR:2020:JA-POLS,Erez:2022:ECCV:MCBM}, and temporal synchronization of multiple videos~\cite{Naaman:ICCV:2025:MultiVideo}. STNs have also been combined with GANs~\cite{Goodfellow:NIPPS:2014:GAN} to generate high-quality canonical atlases~\cite{Mu:CVPR:s022:coordgan,Peebles:CVPR:2022:gangealing}, albeit data demanding. 

Recent works have increasingly adopted deep features as the basis for JA. DINO features~\cite{Caron:ICCV:2021:VITDINO}, in particular, offer robust and semantically-rich representations well-suited for this task. Neural Congealing~\cite{Ofri:CVPR:2023:neuracongealingl} employs test-time optimization  to build class-specific atlases (e.g., birds) by aligning DINO features using rigid and non-rigid warps, predicted by a ResNet-based STN~\cite{He:ECCV:2016:resnet}. ASIC~\cite{Gupta:ICCV:2023:ASIC}  utilizes DINO features, learning dense warps from input images to a canonical space through a U-Net architecture~\cite{Ronneberger:MICCAI:2015:Unet}. Both these methods are computationally intensive (often exceeding an hour for 30 images on an RTX 4090), require heavy regularization to avoid degenerate solutions, and are prone to instability. ASIC also typically produces fragmented or globally-incoherent alignments due to the challenges of dense warping. More recently, a previous work from our group introduced SpaceJAM~\cite{Barel:ECCV:2024:spacejam}, a more efficient solution using a lightweight ConvNet (CNN) and an inverse-compositional loss over refined DINO features, reducing runtime to a few minutes. However, its reliance on high-dimensional feature maps results in substantial memory overhead during the optimization process. In contrast, FastJAM combines sparse KP-based matching with a geometric loss, enabling much faster optimization and better scalability.
FastJAM also differs from SpaceJAM in its architecture and the fact
that the input to its neural net is based on the entire image collection,
as opposed to SpaceJAM's single-image input. 

\paragraph{JA by KP correspondence.}
Several methods tackle JA by explicitly leveraging KP correspondences across the image set and minimizing a geometric loss. Shokrollahi~\etal~\cite{Shokrollahi:CVPR:2015:aligngraph} construct a similarity graph from KP matches to select optimal references for alignment. Safdarnejad~\etal~\cite{Safdarnejad:ECCV:2016:temporally} propose a temporally-aware congealing method for video frames based on tracked KPs. FlowWeb~\cite{zhou:CVPR:2015:flowweb} estimates dense correspondences across images and refines them jointly through graph-based optimization. ASIC~\cite{Gupta:ICCV:2023:ASIC} also incorporates an initial KP matching step, though its primary pipeline relies on dense warping.
Dense matching is often computationally demanding and memory intensive, while sparse KPs offer a more efficient alternative. Building on recent successes of GNNs in pairwise alignment tasks~\cite{Sarlin:CVPR:2020:superglue,Lindenberger:ICCV:2023lightglue, Gava:2023:CVPR:sphereglue}, FastJAM capitalizes on this sparsity and introduces a GNN to aggregate alignment information across matched KPs. Combined with a fast inverse-compositional loss, this enables high-quality and regularization-free JA in seconds and with a low-memory footprint. 
\section{Method}\label{Sec:Method}
\begin{figure}
    \centering
    \includegraphics[width=0.9\linewidth]{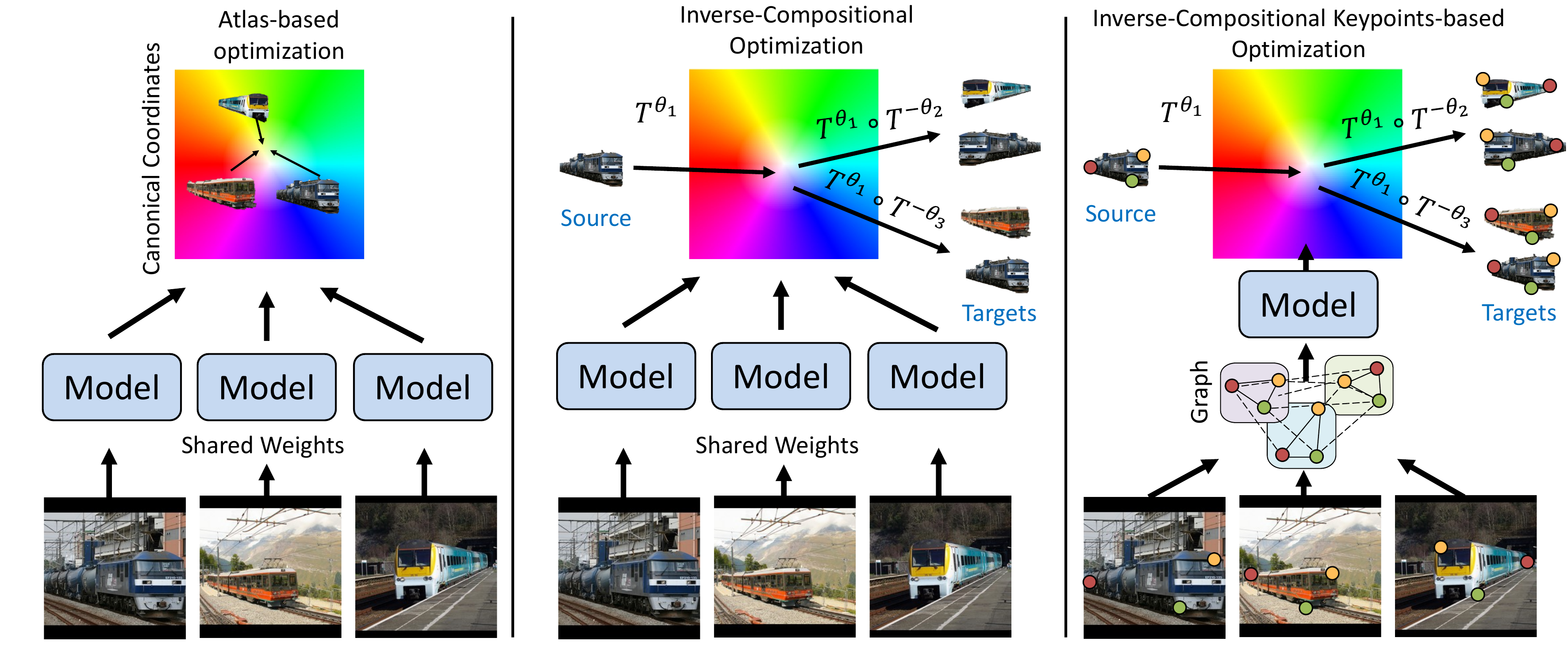}
    \caption{\textbf{Comparison of joint alignment frameworks.} \textit{Left:} Atlas-based methods align each image independently to a canonical space $\mathcal{C}$ by minimizing variance. \textit{Middle:} Existing inverse-compositional (IC) methods estimate $\mathcal{C}$ implicitly via relative transformations ($T^{\btheta_i} \circ T^{-\btheta_j}$), but process images independently. \textit{Right:} \textbf{FastJAM} follows the IC paradigm, but differs from previous approaches in that 1) the loss is computed between KPs and 2) all images are processed simultaneously (the model process the entire KPs graph during its forward pass), allowing shared reasoning across all images.
    }
    \label{fig:framework:comparison}
\end{figure}
In this section, we first formally introduce the JA problem (\autoref{method:subsec:JA:problem}). We then detail how to construct a correspondence graph from pairwise image matches (\autoref{method:subsec:graph:construction}), while in~\autoref{method:subsec:model} we detail the model architecture. Finally, we explain in~\autoref{method:subsec:JA} how to perform image JA with FastJAM.

\subsection{The Joint Alignment Problem and the Inverse-Compositional Framework}\label{method:subsec:JA:problem}
Given  $N$ images, $\Ical=(I_{i})_{i=1}^{N}$, depicting different instances from the same semantic class (\eg, cars), 
the task is to facilitate JA by estimating a transformation $T^{\btheta_i} \in \Tcal$ for each image such that the transformed images $(I_i \circ T^{\btheta_i})_{i=1}^N$ are spatially aligned in a shared coordinate frame $\mathcal{C}$. We assume a parametric family of transformations $\Tcal$ (\eg, homographies), with $T^{\btheta_i}$ denoting the transformation for image $I_i$, parameterized by $\btheta_i$. 
Atlas-based approaches (\eg,~\cite{Ofri:CVPR:2023:neuracongealingl,Gupta:ICCV:2023:ASIC}) usually optimize for a latent template, $I_\mu$, jointly with the transformations.  Formally, they solve 
\begin{align}\label{eq:JA}
    \argmin{I_\mu,(T^{\btheta_i})_{i=1}^N\in\Tcal } \sum\nolimits_{i=1}^N 
     D(I_\mu , I_i\circ T^{\btheta_i})+ \Rcal(T^{\btheta_i};\lambda)\,    
\end{align}
where $D$ is a discrepancy measure (\eg, the Euclidean distance), $\Rcal(T^{\btheta_i};\lambda)$ is a regularization term on the predicted transformations with HPs $\lambda$,  and $I_\mu$ is the so-called canonical space or atlas. Due to its notion of centrality, 
$I_\mu$ is also known as the average or centroid image. 

In contrast, FastJAM follows a congealing-inspired approach~\cite{Miller:CVPR:2000:learning,Learned:PAMI:2006:align} (particularly, Least-Squares (LS) Congealing~\cite{Cox:CVPR:2008:LS,Cox:ICCV:2009:LS}) which avoids the need to maintain an explicit reference image,
together with the modern Inverse Compositional (IC)
approach we proposed in~\cite{Barel:ECCV:2024:spacejam}. 
 Concretely, the IC approach can be defined via the following loss:
\begin{align}\label{eq:loss:congealing}
   \argmin{(T^{\btheta_i})_{i=1}^N\in\Tcal } \sum\nolimits_{i=1}^{N}
   \sum\nolimits_{j:j\neq i} D(I_j,  I_i \circ T^{\btheta_i}
   \circ
   T^{-\btheta{_j}})  
    \,.
\end{align}
Of note, the historical roots of the IC approach go back to 
the pre-DL era~\cite{Cox:ICCV:2009:LS}. SpaceJAM~\cite{Barel:ECCV:2024:spacejam}, however, rather than 
optimizing over the warping of a single image at a time (as was done in traditional LS-congealing) simultaneously optimizes over all transformations.

The IC formulation indicates that the image collection is mapped to a shared space, since
\begin{align}\label{eq:forward:inverse}
   I_j  \approx
   I_i  \circ T^{\btheta_i}
   \circ
   T^{-\btheta{_j}}
 \Leftrightarrow 
   I_j \circ T^{\btheta{_j}}\approx I_i  \circ T^{\btheta_i}  
  \, .
\end{align}
As explained in~\cite{Barel:ECCV:2024:spacejam},
the IC approach obviates the need
for using regularization terms. 
An important distinction between our work and~\cite{Barel:ECCV:2024:spacejam} is that in~\cite{Barel:ECCV:2024:spacejam} the optimization is based on the discrepancy between dense feature maps (\eg, DINO features), while we adapt it to KPs and rely on geometric measure, as detailed below. \autoref{fig:framework:comparison} illustrates the different JA approaches. 

By design, and due to the inverse-compositional nature, IC losses are invariant to a single global homography. That is, for any $(T^{\btheta_i})_{i=1}^N$
and any additional transformation $T^{\btheta_{0}}$, 
the $( T^{\btheta_i}\circ T^{\btheta_0})_{i=1}^N$ transformations
would give rise to the same value of the loss. 
The same phenomena happens in not only~\cite{Barel:ECCV:2024:spacejam}
(and, for slightly-different reasons,~\cite{Shapira:ICML:2023:RFDTAN,Naaman:ICCV:2025:MultiVideo})
but also many works on synchronization over groups (see, \eg, ~\cite{Rosen:IJRS:2019:SESync,Arrigoni:IJCV:2020:Sync}).
This is a feature, not a bug, as it simplifies the optimization considerably. Importantly:
1) While this means there are infinitely-many solutions, 
the implied $N$-fold joint correspondence is unique. 
2) After the fact (\ie, after the optimization is done), 
and for visualization purposes, for example, one can pick the value of
$T^{\btheta_0}$ without affecting the quality of the solution. 
Plausible choices include the inverse of the average homography (which can be computed in various ways, including the so-called Karcher mean~\cite{Pennec:NSIP:1999,karcher2014riemannian}) or the inverse of, say, $T^{\btheta_1}$. Note that the latter choice does \emph{not} imply
that the first image had any special importance in the optimization.

\subsection{Graph Construction}\label{method:subsec:graph:construction}
\begin{figure}[t]
    \centering
    \includegraphics[trim=65mm 20mm 65mm 20mm, clip, width=0.9\linewidth]{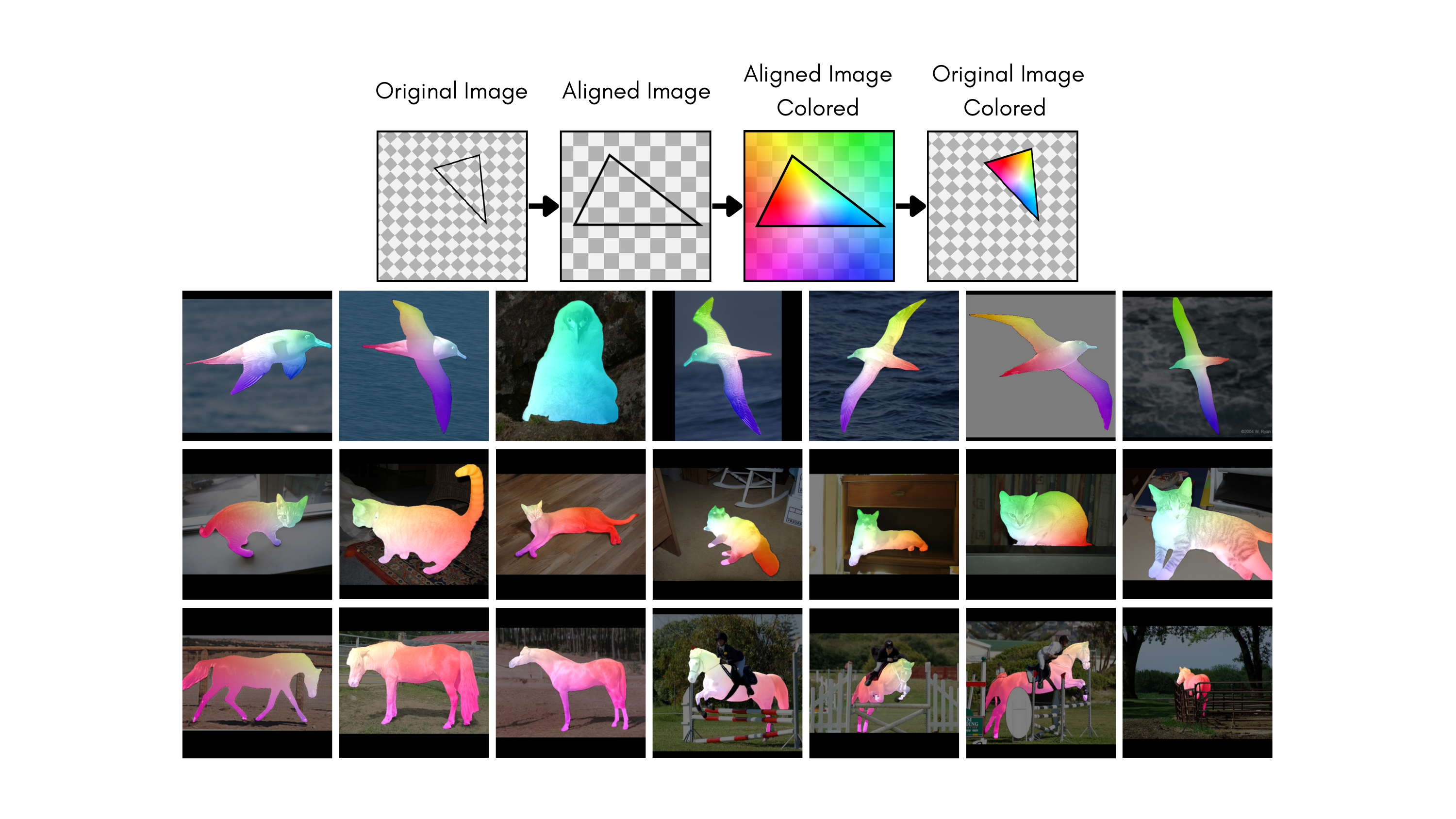}
    \caption{\textbf{Canonical Space Visualization.} We visualize the canonical space ($\mathcal{C}$) via a predefined RGB colormap. The first row shows an example of color projection from the canonical space onto a reference triangle. From the second row, we color each image $I_i$ by applying its inverse transformation on $\mathcal{C}$ (\ie, $\mathcal{C}\circ T^{-\btheta_i}$). FastJAM maps semantically similar regions to the same areas of $\mathcal{C}$, as shown by the consistent color mapping. 
    }
    \label{Fig:Colormap}
\end{figure}
The first stage of FastJAM involves constructing a sparse graph over KPs extracted from the image collection. This graph encodes both intra-image structure and inter-image correspondences and serves as the input to the GNN. The construction process consists of three steps,
detailed below.

\paragraph{Object-centric region extraction.}
We follow~\cite{Ofri:CVPR:2023:neuracongealingl,Gupta:ICCV:2023:ASIC,Barel:ECCV:2024:spacejam} and extract object-centric masks for the image collection. 
We use Grounded-SAM~\cite{Ren:arXiv:2024:GroundedSAM}, a combination of the Grounding DINO object detector and the Segment Anything Model (SAM)~\cite{Caron:ICCV:2021:VITDINO,Kirillov:ICCV:2023:SAM}. Given a text prompt corresponding to the object category, Grounded-SAM produces a segmentation mask focusing on the object of interest. This mask is used to restrict the KP extraction and matching to the relevant region, improving robustness to background clutter and occlusions. Mask extraction takes $\sim0.4$ seconds per image and can be parallelized over the GPU.

\paragraph{KP detection and matching.}
For each image $I_i$, we extract a set of sparse KPs $X_i = \{x_i^{(1)}, \dots, x_i^{(M_i)}\}$ (where $M_i$ is the number KPs in $I_i$)  using an off-the-shelf image matcher inside the objects' mask. We use RoMa~\cite{Edstedt:CVPR:2024:RoMa}, but any matcher producing KP correspondences can be used. We chose RoMA due to its robustness and fast inference time ($\sim0.3$ seconds for an image pair on an RTX4090, which can be done in parallel across pairs). 
For each image pair $(I_i, I_j)$, RoMa returns a set of KPs, $(X_i, X_j)$, and correspondences $\mathcal{M}_{ij} \subset X_i \times X_j$ along with a confidence score for each match. To improve spatial coverage and reduce redundancy, we apply non-maximum suppression (NMS) over the matcher confidence scores using a $30\times30$ window, and retain the top-scoring points. We found that selecting as few as 10 KPs per image is sufficient for our framework.

\paragraph{Intra-image KP clustering.}  
Consider three images, $(I_{1}, I_{2}, I_{3})$, and recall that RoMa is a \textit{pairwise} KP extractor and matcher. Running RoMa on $(I_{1}, I_{2})$ produces 10 KPs on $I_{1}$ and 10 on $I_{2}$; running it on $(I_{1}, I_{3})$ produces another 10 KPs on $I_{1}$ and 10 on $I_{3}$; and so on. Consequently, each image (e.g., $I_{1}$) accumulates multiple sets of KPs obtained from different pairwise runs.  
Importantly, these sets are not guaranteed to be identical, even though they all refer to the same image, leading to redundancy. For instance, in bird images, most of the $N{-}1$ KP sets extracted from $I_{1}$ are likely to include a KP near the beak tip, though at slightly different locations.  
To reduce this redundancy and merge semantically similar KPs within an image, we apply a fast, nonparametric clustering step. Specifically, we use Dinari and Freifeld's parallel DP-Means algorithm~\cite{Dinari:PMLR:2022:dDPMeans} (with \texttt{init\_n}=3 and $\delta=1$), a highly efficient variant of DP-Means~\cite{Kulis:ICML:2011:DPMeans}, which itself generalizes $K$-Means to an unknown number of clusters. It takes $\sim0.13$ seconds to cluster $\sim6000$ points. After clustering, we discard the original KPs and retain only the cluster means as the representative intra-image KPs.

\paragraph{Graph definition.}
We define a single graph $\Gcal = (\Vcal, \Ecal)$ over the entire image set. Each KP, $x_i^{(m)}$, is represented as a node $v \in \Vcal$. We add two types of edges:
    (1) \textbf{Intra-image edges}: In each image $I_i$, we fully connect all KPs in $X_i$ to model local spatial structure.
    (2) \textbf{Inter-image edges}: In each matched KP pair $(x_i^{(m)}, x_j^{(n)}) \in \mathcal{M}_{ij}$, we add an edge between the corresponding nodes.
Each node $v \in \Vcal$ is initialized with a vector $\bh_v^{(0)}$ consisting of the KP's 2D coordinates such that $\bH^{(0)}$ is the initial nodes coordinates matrix. In addition, each node is tagged with a categorical identifier indicating its source image, which is later used to perform image-level pooling. Unlike traditional graph-level readout layers that summarize the entire graph, FastJAM performs structured readout by pooling node embeddings per image.
The edge structure is encoded as a binary adjacency matrix $\bA \in \{0,1\}^{|\Vcal| \times |\Vcal|}$, where $\bA_{uv} = 1$ if there is an edge (intra- or inter-image) between nodes $u$ and $v$. The resulting graph $\Gcal = (\Vcal, \Ecal, \bH^{(0)}, \bA)$ encodes both local geometric structure and cross-image semantic correspondence, and serves as input to a message-passing GNN that propagates alignment information throughout the image collection.

\subsection{Model Architecture}\label{method:subsec:model}
Given the graph $\Gcal = (\Vcal, \Ecal, \bH^{(0)}, \bA)$, our objective is to predict a transformation parameter vector $\btheta_i \in \RR^8$ for each image $I_i$, representing its homography.
We treat this as a structured regression task over node features: each node corresponds to a KP, and the GNN must propagate alignment-relevant information across the graph to produce per-image outputs. 
Formally, let $f(\Gcal)=(\btheta_i)^{N}_{i=1}$ be a GNN that predicts the warping parameters from the graph. 
In our setting, where the goal is to regress image-specific transformations from KPs structured within a shared graph, it is essential to preserve the distinction between a node and its neighbors. GraphSAGE~\cite{Hamilton:NIPS:2017:Sage} achieves this by applying separate transformations to self and neighbor features, enabling more effective modeling of local asymmetries and node-specific roles which is relevant when propagating alignment cues across inter- and intra-image connections. While attention-based models such as GAT~\cite{Velickovic:2017:GAT} offer expressive edge-aware aggregation, we found them to be empirically slower. In contrast, GraphSAGE provided a favorable balance of speed, stability, and alignment accuracy, making it well-suited for FastJAM’s test-time optimization regime. The message protocol for GraphSAGE is 
\begin{align}\label{eq:graphsage}
    \bh_v^{(l)} = \sigma\left(\bW_1^{(l)}\bh_v^{(l-1)} + \bW_2^{(l)}\cdot\mathrm{mean}_{u \in \mathcal{N}(v)}  \bh_u^{(l-1)} \right)
\end{align}

where $\bh_v^{(l)}$ is the embedding of node $v$ at layer $l$, $\mathcal{N}(v)$ denotes the neighbors of $v$, $\bW^{(l)}$ is a learnable weight matrix, and $\sigma$ is a non-linear activation function.
After $L=5$ message-passing layers, we perform \textbf{per-image readout} via global average pooling over all nodes belonging to each image:
\begin{align}
    \bz_i = \tfrac{1}{M_i} \sum\nolimits_{v \in \Vcal_i} \bh_v^{(L)}
\end{align}
where $\Vcal_i \subset \Vcal$ is the set of nodes from image $I_i$, and $\bz_i \in \RR^d$ is the resulting image-level embedding. Finally, we project the embedding of each image to the estimated homography parameters:

\subsection{FastJAM Joint Alignment}\label{method:subsec:JA}

\paragraph{Lie-algebraic parameterization.}

To ensure matrix invertibility, which is essential for our IC formulation and for stable optimization~\cite{Skafte:CVPR:2018:DDTN,Barel:ECCV:2024:spacejam}, we represent homographies using the Special Linear group $\mathrm{SL}(3)$ via a Lie-algebraic parameterization as was done in, \eg,~\cite{mei2006homography,Arrigoni:IJCV:2020:Sync,Barel:ECCV:2024:spacejam}. 
For details, see our supplemental material (\textbf{SupMat}).

\paragraph{Robust inverse-compositional KP loss.}
Our geometric loss builds upon the IC formulation introduced in~\cite{Barel:ECCV:2024:spacejam}, adapted to sparse KP correspondences. For each image pair $(I_i, I_j)$, the forward warp from $I_i$ is composed with the inverse warp from $I_j$ (\ie, $ T^{\btheta_i} \circ T^{-\btheta_j}$). We penalize the discrepancy between each matched KP pair $(x_i^{(m)}, x_j^{(n)}) \in \mathcal{M}_{ij}$ after applying the IC transformation

\begin{align}\label{eq:kpic:loss}
    \mathcal{L}_{\text{KP-IC}} = \sum_{i=1}^{N} \sum_{j \ne i} \sum_{(x^{(n)}_i, x^{(m)}_j) \in \mathcal{M}_{ij}} \rho_\sigma\left( \lVert x^{(m)}_j - x^{(n)}_i\circ T^{\btheta_i} \circ T^{-\btheta_j} \rVert_2 \right),
\end{align}
where $\rho_\sigma(z) = \frac{z^2}{z^2 + \sigma^2}$ is the Geman-McClure robust loss function~\cite{GemanMcClure:BISI:1987} with parameter $\sigma$.

This formulation allows alignment to be computed at the original KP locations without any regularization term on the warps or the need to render warped images via expensive interpolation. Compared to dense alignment over high-dimensional DINO feature maps, our sparse formulation is both significantly more efficient and more robust to missing KPs, wrong matches, and outliers.

\paragraph{Handling reflections.}
We follow~\cite{Barel:ECCV:2024:spacejam} and explicitly check for flips every $K$ epochs during optimization (where $K=100$) and compute the gradient and update the model's weight only for the best configuration. We have found that only checking for horizontal flips is sufficient. This ensures that flipped images can still participate in alignment without requiring a reflection-aware parameterization.

\paragraph{Implementation details.}
All experiments were conducted on a single NVIDIA RTX 4090 GPU with 24GB of memory. We optimize FastJAM for 600 epochs using Adam~\cite{Kingma:arxiv:2014:Adam} with a Geman-McClure robustness parameter $\sigma = 0.25$. We use pretrained Grounding-SAM~\cite{Ren:arXiv:2024:GroundedSAM} and RoMa~\cite{Edstedt:CVPR:2024:RoMa} with the default HP once, before starting the optimization. For more details, please see our \textbf{SupMat}. 

\paragraph{Limitations.}\label{limitations} Our main limitation is the reliance on an external image matcher to generate initial KP correspondences. While modern matchers like RoMa provide high-quality matches in many scenarios, the overall alignment quality depends on the accuracy of these correspondences. In addition, FastJAM models geometric transformations using homographies, which may be insufficient in cases involving strong non-planar deformations. Extending the model to support more expressive transformation families remains a direction for future work.

\section{Results}\label{Sec:Results}
\paragraph{Datasets, evaluation metrics, and baselines.}
We evaluate FastJAM 
under a test-time optimization setting, where the model is optimized independently on each image collection. We use two benchmark datasets: \textbf{SPair-71k}~\cite{Min:2019:spair} and \textbf{CUB-200}~\cite{Wah:2011:CUB} (classes and subsets). SPair-71k's test set 
comprises 18 object categories, each with $\sim$30 images, with annotated KPs and large intra-class variation. We report both per-category performance and average results across all categories.
Following prior works, we evaluate on the first 3 categories of CUB-200 test set, each containing $\sim$30 images as well.
We use the \textbf{Percentage of Correct Keypoints (PCK)} as the evaluation metric. A predicted KP is deemed correct if it falls within a normalized distance threshold $\alpha$ of the ground-truth location. We report mean PCK across all KPs and categories, with $\alpha=0.1$. We average the results over 3 runs. 

As in prior work~\cite{Gupta:ICCV:2023:ASIC,Barel:ECCV:2024:spacejam}, we compare FastJAM to several baselines. Neural Best Buddies (NBB)~\cite{Aberman:TOG:2018:bestbuddies} aligns image pairs using mutual nearest neighbors with Moving Least Squares warping~\cite{Schaefer:ACM:2006:MLS}, using VGG (\textbf{VGG-MLS}) or DINO (\textbf{DINO-MLS}) features. \textbf{DINO-NN} performs dense nearest-neighbor matching. \textbf{GANgealing}~\cite{Peebles:CVPR:2022:gangealing} uses GANs but is limited to seen categories. \textbf{Neural Congealing}~\cite{Ofri:CVPR:2023:neuracongealingl} builds an explicit atlas but requires hyperparameter tuning and  reported results on only three SPair-71k classes. \textbf{ASIC}~\cite{Gupta:ICCV:2023:ASIC} predicts dense warps to a canonical space, and \textbf{SpaceJAM}~\cite{Barel:ECCV:2024:spacejam} applies an IC loss over DINO features. FastJAM instead uses sparse KPs and a graph-based model, enabling faster and more scalable alignment.
We cite the results reported in~\cite{Gupta:ICCV:2023:ASIC,Barel:ECCV:2024:spacejam}.

\subsection{Qualitative Results}\label{results:subsec:qualitative}
\begin{figure}[t]
    \centering
    \includegraphics[trim=70mm 15mm 75mm 15mm, clip, width=0.85\linewidth]{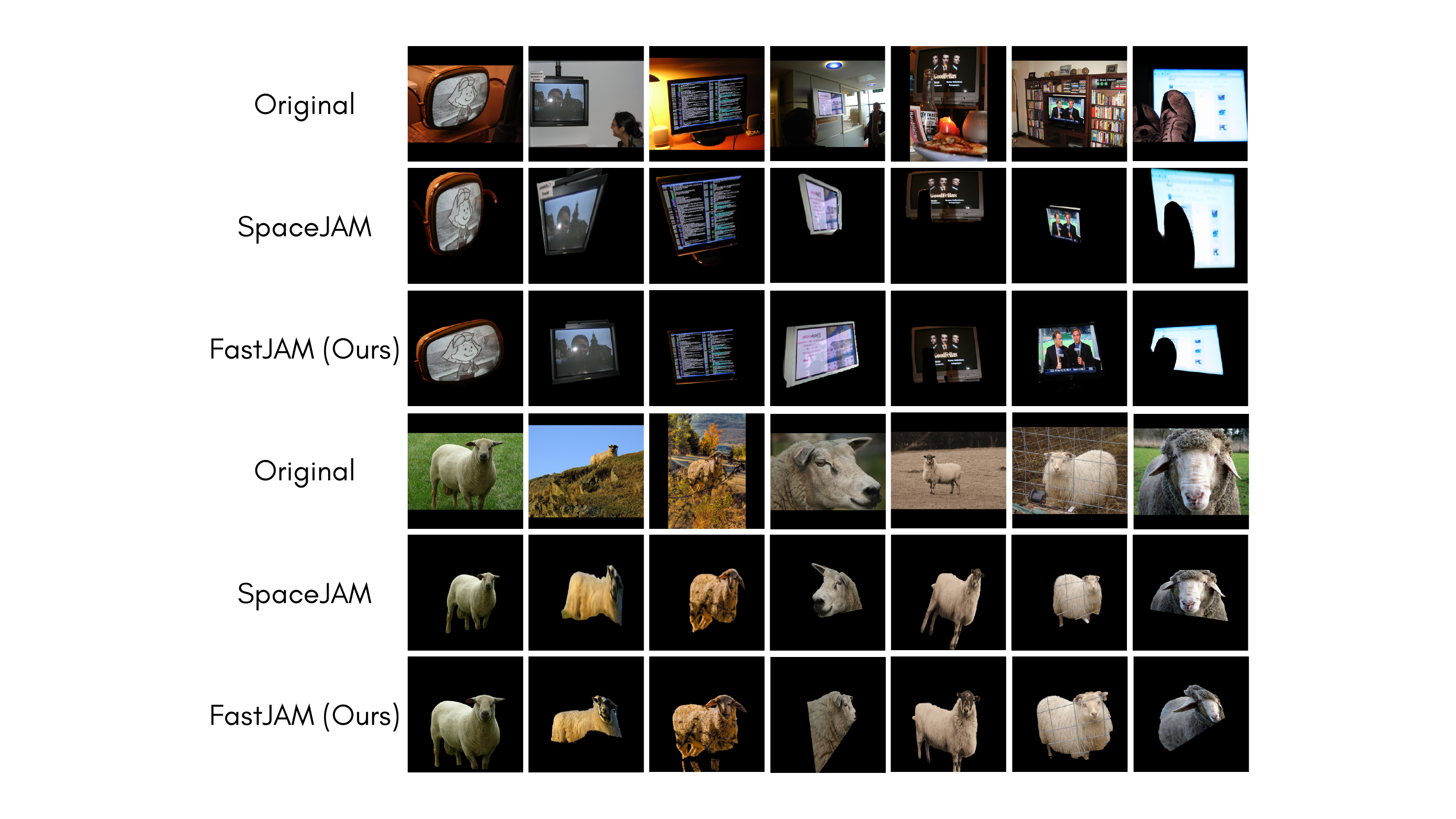}
    \caption{\textbf{JA Visual Comparison.} We compare FastJAM   with SpaceJAM~\cite{Barel:ECCV:2024:spacejam} on both rigid (TV) and non-rigid (Sheep) classes. In both cases, FastJAM alignment is visually better, where the improvement is particularly noticeable for close-up images, such as the middle or rightmost sheep.} 
    \label{Fig:Comp}
\end{figure}

We illustrate the qualitative performance of FastJAM in \autoref{Fig:Alig}, \autoref{Fig:Colormap}, and \autoref{Fig:Comp}.  
Given a collection of category-level images (\eg, birds), FastJAM aligns all images within seconds, producing visually coherent and semantically consistent outputs across instances.  
To interpret the alignment, we visualize the canonical space $\mathcal{C}$ using a fixed RGB colormap. Each image $I_i$ is colored by applying the inverse of its predicted transformation to $\mathcal{C}$, \ie, $\mathcal{C} \circ T^{-\btheta_i}$.  
As shown in \autoref{Fig:Colormap}, semantically similar regions (\eg, heads, wings, tails) align to consistent areas in $\mathcal{C}$, indicating robust JA across pose and appearance. As shown in \autoref{Fig:Comp} in comparison to~\cite{Barel:ECCV:2024:spacejam}, FastJAM alignment is visually better. Additional examples and comparisons are available in the \textbf{SupMat}.

\subsection{Quantitative Results}\label{results:subsec:quantitative}
\autoref{tab:spair:full} reports alignment accuracy on the SPair-71k test set, measured by mean PCK@0.10 across 18 object categories. FastJAM achieves the best overall performance with an average PCK of \textbf{53.0}, outperforming all competing methods. It ranks first in 11 categories and second in 6 others, with comparable performance across a wide range of object classes and viewpoints.
Compared to~\cite{Ofri:CVPR:2023:neuracongealingl,Gupta:ICCV:2023:ASIC}, FastJAM offers on-par or better accuracy while being significantly faster and more memory-efficient, validating the benefits of its sparse, graph-based formulation.
While FastJAM performs competitively across most categories, we observe reduced performance on highly-symmetric objects such as \textit{bicycles}. These cases pose inherent challenges due to visual ambiguity, where the initial matcher struggles to disambiguate symmetric parts (\eg., right versus left). In such scenarios, FastJAM can propagate incorrect correspondences. Addressing this remains an interesting direction for future work.
Results for the CUB-200 dataset are reported in~\autoref{tab:CUB} where FastJAM outperforms SpaceJAM and achieves comparable results to ASIC~\cite{Gupta:ICCV:2023:ASIC} (in a fraction of the computation time) on the categories benchmark and the best results across subsets (results were not reported for~\cite{Gupta:ICCV:2023:ASIC}). 

\begin{table}[t]
\centering
\caption{\textbf{SPair-71k results:} PCK@$0.10$ on the test set. Among test-time optimization (TTO) methods, the best is in \textbf{bold}, second-best is \underline{underlined}. ($\star$) Denotes use of a reference image. ($-$) Indicates missing results. (\textdagger) Marks non-TTO methods.}
\label{tab:spair:full}
\resizebox{\textwidth}{!}{
\begin{tabular}{@{}lcccccccccccccccccc|c@{}}
\toprule
\textbf{Method} & 
Aero & Bike & Bird & Boat & Bottle & Bus & Car & Cat & Chair & Cow & Dog & Horse & Motor & Person & Plant & Sheep & Train & TV & \textbf{All} \\  

\midrule

GANgealing~\cite{Peebles:CVPR:2022:gangealing} & 
- & 37.5\textsuperscript{\textdagger} & - & - & - & - & - & 67.0 & - & - & 23.1 & - & - & - & - & - & - & 57.9 & - \\
\hline
VGG+MLS~\cite{Aberman:TOG:2018:bestbuddies} & 
29.5 & 22.7 & 61.9 & {26.5} & 20.6 & 25.4 & 14.1 & 23.7 & 14.2 & 27.6 & 30.0 & 29.1 & 24.7 & 27.4 & 19.1 & 19.3 & 24.4 & 22.6 & 27.4 \\

 DINO+MLS~\cite{Aberman:TOG:2018:bestbuddies,Caron:NIPS:2020:unsupervised} & 
49.7 & 20.9 & 63.9 & 19.1 & 32.5 & 27.6 & 22.4 & 48.9 & 14.0 & 36.9 & 39.0 & 30.1 & 21.7 & 41.1 & 17.1 & 18.1 & 35.9 & 21.4 & 31.1 \\

 DINO+NN~\cite{Amir:2021:VITdescriptoers} & 
{57.2} & 24.1 & \underline{67.4} & 24.5 & 26.8 & 29.0 & 27.1 & 52.1 & {15.7} & {42.4} & {43.3} & 30.1 & 23.2 & 40.7 & 16.6 & 24.1 & 31.0 & \underline{24.9} & 35.0 \\
\hline

 NeuCongeal~\cite{Ofri:CVPR:2023:neuracongealingl} & 
- & {29.1$^\star$} & - & - & - & - & - & 53.3 & - & - & 35.2 & - & - & - & - & - & - & - & - \\

 ASIC~\cite{Gupta:ICCV:2023:ASIC} & 
\underline{57.9} & 
{25.2} & 
\textbf{68.1} & 
{24.7} & 
35.4 & 
28.4 & 
{30.9} & 
{54.8} & 
{21.6} & 
{45.0} & 
\textbf{47.2} & 
{39.9} & 
{26.2} & 
{48.8} & 
14.5 & 
\underline{24.5} & 
{49.0} & 
24.6 &  
{37.0} \\ 

 SpaceJAM (ViT-L)~\cite{Barel:ECCV:2024:spacejam} &
{53.6} & 
{\textbf{53.5}} & 
{45.4} & 
{\textbf{47.5}} & 
{\textbf{71.0}} & 
{\underline{54.0}} & 
{\underline{46.0}} & 
{\textbf{66.0}} & 
{\underline{25.8}} & 
{\underline{48.6}} & 
{28.5} & 
{\underline{47.6}} & 
{\underline{54.0}} & 
{\textbf{50.7}} & 
{\underline{34.0}} & 
{09.0} & 
{\underline{71.8}} & 
{15.4} & 
{\underline{45.7}} \\ 

 FastJAM (Ours)&
{\textbf{64.4}} & 
{\underline{43.3}} & 
{60.0} & 
{\underline{29.6}} & 
{\underline{58.4}} & 
{\textbf{66.8}} & 
{\textbf{56.5}} & 
{\underline{63.7}} & 
{\textbf{32.0}} & 
{\textbf{49.2}} & 
{\underline{40.8}} & 
{\textbf{53.7}} & 
{\textbf{62.8}} & 
{\underline{49.1}} & 
{\textbf{42.9}} & 
{\textbf{33.4}} & 
{\textbf{76.2}} & 
{\textbf{71.2}} & 
{\textbf{53.0}} \\ 

\bottomrule

\end{tabular}}
\end{table}
\def\ASIC{Gupta:ICCV:2023:ASIC}
\def\DINOMLS{Caron:NIPS:2020:unsupervised, Aberman:TOG:2018:bestbuddies}
\def\DINONN{Amir:2021:VITdescriptoers}
\def\VGGMLS{Aberman:TOG:2018:bestbuddies}

\begin{table}[t]
\centering
\resizebox{0.9\textwidth}{!}{
\begin{minipage}[t]{0.55\textwidth}
    \centering
    \captionsetup{justification=centering}
    \tiny
    \caption{A comparison on CUB-200.}
    \label{tab:CUB}
    \begin{tabular}{lc|cc}
    \toprule
    \textbf{Method} & CUB-200 & \textbf{Method} & CUB-200 \\
                    & (first 3 cate.) &               & (Subsets) \\
    \midrule
    VGG+MLS~\cite{\VGGMLS}   & 25.8 & - & - \\
    DINO+MLS~\cite{\DINOMLS} & 67.0 & -  & - \\
    DINO+NN~\cite{\DINONN}   & 68.3 & GANgealing~\cite{Peebles:CVPR:2022:gangealing} &  56.8 \\
    ASIC~\cite{\ASIC}        & \textbf{75.9} & NeuCongeal~\cite{Ofri:CVPR:2023:neuracongealingl} & 63.6 \\
    SpaceJAM~\cite{Barel:ECCV:2024:spacejam}       & 69.6 & SpaceJAM~\cite{Barel:ECCV:2024:spacejam}  & \underline{69.9} \\
    FastJAM (Ours) & \underline{75.3} & FastJAM (Ours)& \textbf{73.6} \\
    \bottomrule
    \end{tabular}
    \vspace*{0.18 cm}
    \centering
    \tiny
    \caption{Dense vs. KP warping runtime analysis [sec]. \\}
    \label{tab:runtime}
    \setlength{\tabcolsep}{2pt}
    \begin{tabular}{lccc|cccc}
    \toprule
    \textbf{Loss} & $B_{\textrm{max}}$ & $N_{\textrm{points}}$ & $D$ & Grid warping & Interpolation  & Fwd. +Back  & Total\\
    \midrule
    Dense & 10 & $70756$ & 2 & 1.28 & 0.13 & $2.82$ & $8.46$ \\
    Dense & 10 & $70756$ & 25 & 1.29 & 0.77 & $4.12$ & $12.36$ \\
    KPs & 30 & 16 & 2 & 0.18 & \noindent\rule{0.5cm}{0.4pt} & $0.36$ &  $0.36$ \\
    KPs & 30 & 16 & 25 & 0.17 & \noindent\rule{0.5cm}{0.4pt} & $0.34$ & $0.34$ \\
    \bottomrule
    \end{tabular}
\end{minipage}
\begin{minipage}[t]{0.4\textwidth}
    \centering
    \tiny
    \caption{Ablation Study.}
    \label{tab:ablation}
    \setlength{\tabcolsep}{3pt}
    \begin{tabular}{lcc}
    \toprule
    \textbf{Ablation} & CUB-200 & SPair-71k \\
                      & (first 3 cate.)\\
    \midrule
    LoFTR \cite{Sun:CVPR:2021:loftr} (No RoMa)      & 33.8     & 17.0 \\
    \midrule
    Linear projection & 33.2 & 14.0 \\
    MLP (per-image, no graph) & 73.5 & 47.8 \\
    Homography optimization (no deep net) & 74.1 & 48.3 \\
    \midrule
    L2 loss (No Geman-McClure function)            & 61.0      & 33.9 \\
    No masks & 74.8      & 41.8 \\
    No non-maximum suppression & 74.3      & 47.2 \\
    No Lie Group & 75.0   & 49.3 \\
    No intra-image edges & 74.7 & 50.0 \\
    \midrule
    GNN Backbones & & \\
    \midrule
    GCN~\cite{Kipf:ICLR2017:GCN} (53 secs)                & 73.2      & 47.4 \\
    GAT~\cite{Velickovic:2017:GAT} (66 secs)                & 75.0      & 49.9 \\
    GraphSAGE~\cite{Hamilton:NIPS:2017:Sage} (49 secs)     & \textbf{75.3} & \textbf{53.0} \\
    \bottomrule
    \end{tabular}
\end{minipage}
}
\end{table}

\subsection{Runtime Analysis}\label{results:subsec:runtime}
\textbf{Runtime comparison.} We evaluate the computational efficiency of FastJAM by comparing its runtime against: NeuCongeal~\cite{Ofri:CVPR:2023:neuracongealingl}; ASIC~\cite{Gupta:ICCV:2023:ASIC}; SpaceJAM~\cite{Barel:ECCV:2024:spacejam}.   
As shown in \autoref{tab:ja:comparison}, FastJAM achieves over an \textbf{order-of-magnitude speedup} (measured over three SPair-71k categories), aligning image collections in  \textbf{under $\sim$50 seconds}, compared to 5–6 minutes for SpaceJAM and over an hour for ASIC and NeuCongeal. The reported runtime includes preprocessing (\ie, pairwise matches). 
We also compare FastJAM and SpaceJAM on an increasing number of images ($N=10$ to $100$). The full experimental setup is available in our \textbf{SupMat}.

\textbf{Dense vs. KP warping.}  
We analyze the individual warping components of both methods on a set of 30 images (see~\autoref{tab:runtime}, all reported runtime in this table are in seconds) for one epoch. We evaluate how the number of points ($N_{\mathrm{points}}$) in the coordinate grid and the feature dimension ($D$) affect the overall warping time.  
For dense matching, we set $N_{\mathrm{points}}=266\times266=70{,}756$ (the image resolution used in SpaceJAM) and $D=25$, corresponding to the feature dimension on which the loss is computed. For FastJAM, we set $N_{\mathrm{points}}=16$ and $D=2$ (\ie, 8 KPs in 2D). For completeness, we also evaluate $D\in\{2,25\}$ for both methods.  
We measure the time required for (\emph{i}) grid warping, (\emph{ii}) interpolation (used only in dense warping), (\emph{iii}) forward and backward warping for the IC loss (\emph{Fwd + Back}), and the total runtime over all batches in a single epoch. For dense warping (\eg, SpaceJAM), the maximum batch size is $B_{\mathrm{max}}=10$, resulting in a total runtime of $3\times(\text{Fwd+Back})$.  
The key observations are: (1) reducing $N_{\mathrm{points}}$ is crucial for achieving fast warping, and (2) avoiding interpolation further accelerates computation, making the runtime largely independent of $D$.

\subsection{Ablation Study}\label{results:subsec:ablation}
\autoref{tab:ablation} summarizes our ablation study on CUB-200 (3 categories) and SPair-71k. Replacing RoMa with LoFTR~\cite{Sun:CVPR:2021:loftr} significantly reduces performance, underscoring LoFTR’s limitations in cross-instance correspondence. Substituting the Geman–McClure loss~\cite{GemanMcClure:BISI:1987} with an $\ell_2$ loss or removing the object mask also causes notable accuracy drops, confirming the importance of robust error modeling and spatial masking. NMS and intra-image edges provide additional gains. Although removing the Lie-algebraic parameterization has little effect on accuracy, it ensures warp invertibility, essential for the IC loss, as without it about 2\% of runs fail due to non-invertible matrices, whereas using it eliminates such failures entirely.

Replacing the GNN with a linear projection caused a substantial performance drop, while MLP-based and direct homography optimization models also reduced accuracy, though less severely. Among GNN backbones, GraphSAGE~\cite{Hamilton:NIPS:2017:Sage} outperforms both GCN~\cite{Kipf:ICLR2017:GCN} and GAT~\cite{Velickovic:2017:GAT} in accuracy and runtime.
\section{Conclusion}
We introduced \textbf{FastJAM}, a graph-based framework for fast and scalable image JA. By leveraging sparse KP correspondences and a lightweight GNN architecture, FastJAM propagates alignment cues across image collections and regresses per-image transformations. Our method achieves state-of-the-art alignment quality while significantly reducing runtime and memory usage.

\clearpage
\section*{Acknowledgments} 
This work was supported by the Lynn and William Frankel Center at BGU CS, 
by the Israeli Council for Higher Education via the BGU Data Science Research Center, 
RSW's work was supported by the Kreitman School of Advanced Graduate Studies. 
Both OH and SI were also supported by the VATAT National excellence scholarship for MSc students in AI and Data Science.
\bibliographystyle{unsrt}
\bibliography{refs}

\clearpage
\newpage
\appendix

 \clearpage
\newpage
\section*{NeurIPS Paper Checklist}

\begin{enumerate}

\item {\bf Claims}
    \item[] Question: Do the main claims made in the abstract and introduction accurately reflect the paper's contributions and scope?
    \item[] Answer: \answerYes{}
    \item[] Justification: All the claims made in the abstract are backed in the paper.

\item {\bf Limitations}
    \item[] Question: Does the paper discuss the limitations of the work performed by the authors?
    \item[] Answer: \answerYes{}
    \item[] Justification: In section ~\autoref{limitations}, we elaborate about the limitations of FastJAM.

\item {\bf Theory assumptions and proofs}
    \item[] Question: For each theoretical result, does the paper provide the full set of assumptions and a complete (and correct) proof?
    \item[] Answer: \answerNA{}
    \item[] Justification: there are no theoretical results in the paper.

    \item {\bf Experimental result reproducibility}
    \item[] Question: Does the paper fully disclose all the information needed to reproduce the main experimental results of the paper to the extent that it affects the main claims and/or conclusions of the paper (regardless of whether the code and data are provided or not)?
    \item[] Answer: \answerYes{}
    \item[] Justification: Most if not all needed information about reproducibility is given either in the paper itself, in the Supplementary or in the future to be published code.

\item {\bf Open access to data and code}
    \item[] Question: Does the paper provide open access to the data and code, with sufficient instructions to faithfully reproduce the main experimental results, as described in supplemental material?
    \item[] Answer: \answerYes{}
    \item[] Justification: As described in the abstract, the code would be released upon acceptance. The datasets we used are publicly available.

\item {\bf Experimental setting/details}
    \item[] Question: Does the paper specify all the training and test details (e.g., data splits, hyperparameters, how they were chosen, type of optimizer, etc.) necessary to understand the results?
    \item[] Answer: \answerYes{}
    \item[] Justification: We describe all the details either in the paper itself or in the supplementary.

\item {\bf Experiment statistical significance}
    \item[] Question: Does the paper report error bars suitably and correctly defined or other appropriate information about the statistical significance of the experiments?
    \item[] Answer: \answerNo{}
    \item[] Justification: Due to the substantial computational cost associated with running all competing methods, it was not feasible to perform multiple trials necessary for estimating variance or reporting error bars. On our results, we averaged our experiments on 3 different seed runs.

\item {\bf Experiments compute resources}
    \item[] Question: For each experiment, does the paper provide sufficient information on the computer resources (type of compute workers, memory, time of execution) needed to reproduce the experiments?
    \item[] Answer: \answerYes{}
    \item[] Justification: The experiments where maybe on internal cluster, GPU RTX4090, we mention it where needed. Memory size is also indicated.
    
\item {\bf Code of ethics}
    \item[] Question: Does the research conducted in the paper conform, in every respect, with the NeurIPS Code of Ethics \url{https://neurips.cc/public/EthicsGuidelines}?
    \item[] Answer: \answerYes{}
    \item[] Justification: All the data we used is publicly available.

\item {\bf Broader impacts}
    \item[] Question: Does the paper discuss both potential positive societal impacts and negative societal impacts of the work performed?
    \item[] Answer: \answerNo{}
    \item[] Justification:  The method presented is a generic tool for aligning images and is not directly tied to a specific application. While it may enable both beneficial and potentially harmful uses (e.g., in surveillance or misinformation), the paper does not explore these societal implications.
    
\item {\bf Safeguards}
    \item[] Question: Does the paper describe safeguards that have been put in place for responsible release of data or models that have a high risk for misuse (e.g., pretrained language models, image generators, or scraped datasets)?
    \item[] Answer: \answerNA{}
    \item[] Justification: We did not found high risk misuse for our method.

\item {\bf Licenses for existing assets}
    \item[] Question: Are the creators or original owners of assets (e.g., code, data, models), used in the paper, properly credited and are the license and terms of use explicitly mentioned and properly respected?
    \item[] Answer: \answerYes{}
    \item[] Justification: Any credit or citation needed was provided in the paper.

\item {\bf New assets}
    \item[] Question: Are new assets introduced in the paper well documented and is the documentation provided alongside the assets?
    \item[] Answer: \answerYes{}
    \item[] Justification: All needed documentation would be provided with the code upon acceptance.

\item {\bf Crowdsourcing and research with human subjects}
    \item[] Question: For crowdsourcing experiments and research with human subjects, does the paper include the full text of instructions given to participants and screenshots, if applicable, as well as details about compensation (if any)? 
    \item[] Answer: \answerNA{}
    \item[] Justification: the paper did not involve crowd-sourcing nor research with human subjects.

\item {\bf Institutional review board (IRB) approvals or equivalent for research with human subjects}
    \item[] Question: Does the paper describe potential risks incurred by study participants, whether such risks were disclosed to the subjects, and whether Institutional Review Board (IRB) approvals (or an equivalent approval/review based on the requirements of your country or institution) were obtained?
    \item[] Answer: \answerNA{}
    \item[] Justification: The paper does not involve crowd-sourcing nor research with human subjects.

\item {\bf Declaration of LLM usage}
    \item[] Question: Does the paper describe the usage of LLMs if it is an important, original, or non-standard component of the core methods in this research? Note that if the LLM is used only for writing, editing, or formatting purposes and does not impact the core methodology, scientific rigorousness, or originality of the research, declaration is not required.
    \item[] Answer: \answerNA{}
    \item[] Justification: The usage of LLM, as declared, was mainly for writing and editing.

\end{enumerate}

\newpage
\pagestyle{empty}


\section*{Supplementary material}

# FastJAM: a Fast Joint Alignment Model for Images

## Supplemental Material

## Contents

This document contains the following:

- **§ A**: **Visual Comparison**
  - Additional qualitative comparisons against existing joint alignment methods (Figure 6).
- **§ B**: **Additional Visualizations**
  - Additional joint alignment results on the CUB dataset (Figure 7).
  - Full pairwise alignment grids for representative categories (Figure 8 and Figure 9).
- **§ C**: **Explaining the Colormap Visualization in More Detail**
  - Illustration of FastJAM's canonical alignment using a fixed RGB colormap (Figure 10).
- **§ D**: **Runtime Analysis**
  - Scalability runtime assessment for increasing number of images ($N = 10$ to $100$) (Figure 11).
- **§ E**: **Model Configuration and Training Setup**
  - Full architecture summary (Table 5).
  - Optimization settings, matcher configuration, and training procedure.
- **§ F**: **External Tools and Frameworks**
  - Summary of third-party tools and libraries used, including RoMa, LoFTR, Grounded-SAM, and `torch_geometric`.
- **§ G**: **Lie Group Parameterization**
  - Full explanation of the Lie group parameterization.

Additionally, key notational conventions used throughout the main paper and this document are summarized in Table 6.

## A  Additional Visual Comparisons

To qualitatively assess alignment quality, we compare FastJAM against existing joint alignment methods, including SpaceJAM and ASIC. As illustrated in Figure 6, FastJAM produces more coherent and natural-looking alignments, particularly in challenging cases involving pose variation. Unlike ASIC, which applies dense warping and often introduces distortions, FastJAM preserves global structure by relying on sparse keypoints and homographic transformations.

## B  Additional Visualizations

This section provides supplementary qualitative results that further demonstrate the alignment capabilities of FastJAM across various categories and settings.

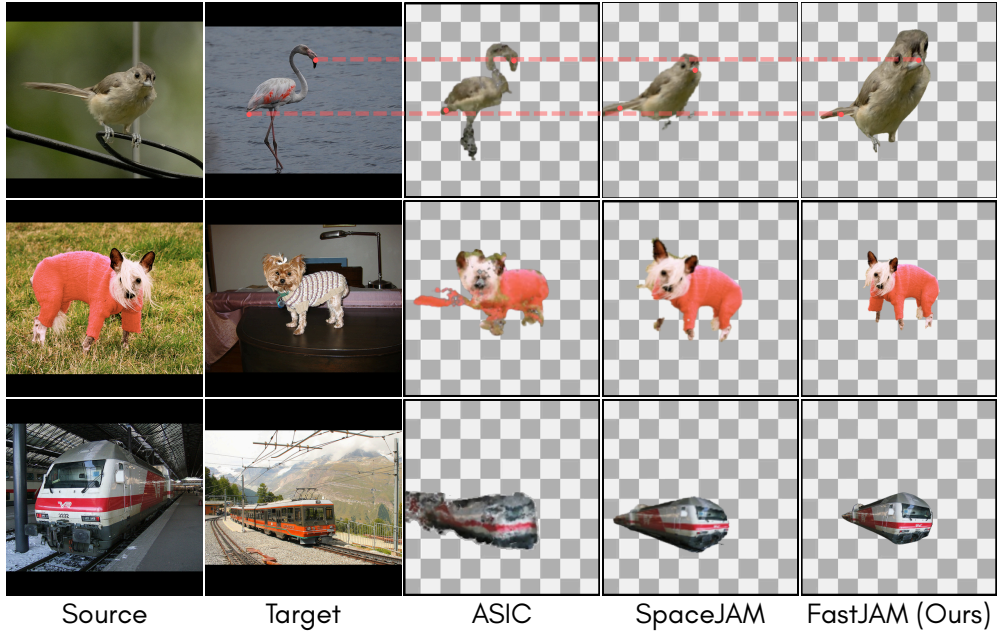

Figure 6: **Qualitative Comparison of Pairwise Alignment Methods.** Visual comparison of joint alignment results across ASIC [1], SpaceJAM [2], and FastJAM (ours) on several image pairs. ASIC, which relies on dense warping fields, often introduces spatial distortions and unrealistic deformations in low-texture or structured regions. In contrast, SpaceJAM and FastJAM apply global homographies, resulting in more coherent transformations. Notably, in the top row (bird), FastJAM produces the most geometrically consistent alignment, as evidenced by the parallel lines and precise correspondence of semantically meaningful points such as the beak and tail.

### B.1 Additional Joint Alignment on CUB Dataset

Figure 7 presents further joint alignment results on additional classes from the CUB-200 dataset. For each class, the original input images are shown in the top row, while the bottom row displays their aligned counterparts. The outputs demonstrate FastJAM's ability to handle fine-grained categories and produce visually coherent canonical views across varying poses and appearances.

### B.2 Pairwise Alignment

We visualize full pairwise alignment grids for two representative categories: "aeroplane" and "horse." As shown in Figure 8 and Figure 9, each grid displays how the source images (rows) are aligned to the target images (columns) using the estimated inverse-compositional warps. The diagonal entries, which correspond to self-alignments (i.e., identity transformations), are highlighted with a purple dashed-line frame. These grids illustrate FastJAM's ability to produce consistent, symmetric mappings across image pairs and maintain semantic structure throughout the alignment process.

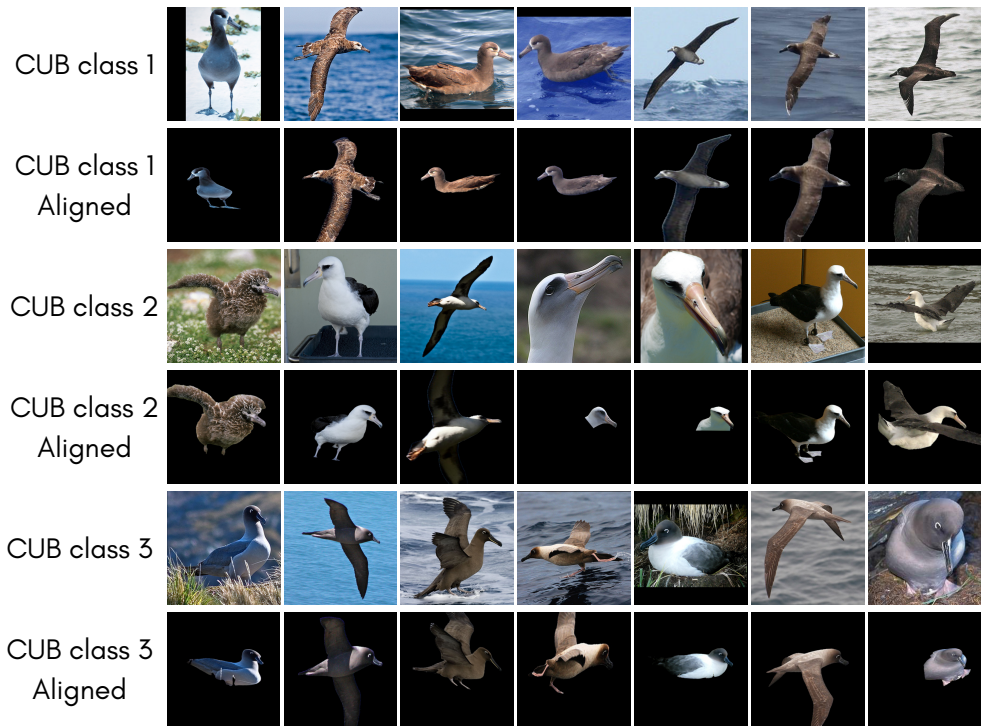

Figure 7: **Qualitative alignment results on CUB-200 classes.** Each pair of rows corresponds to a distinct semantic class from the CUB-200 [3] dataset. The top row in each pair shows the original, unaligned images; the bottom row shows the corresponding aligned images produced by FastJAM. The alignment process successfully maps semantically consistent parts (e.g., heads, wings, tails) to similar spatial locations across different instances within each class.

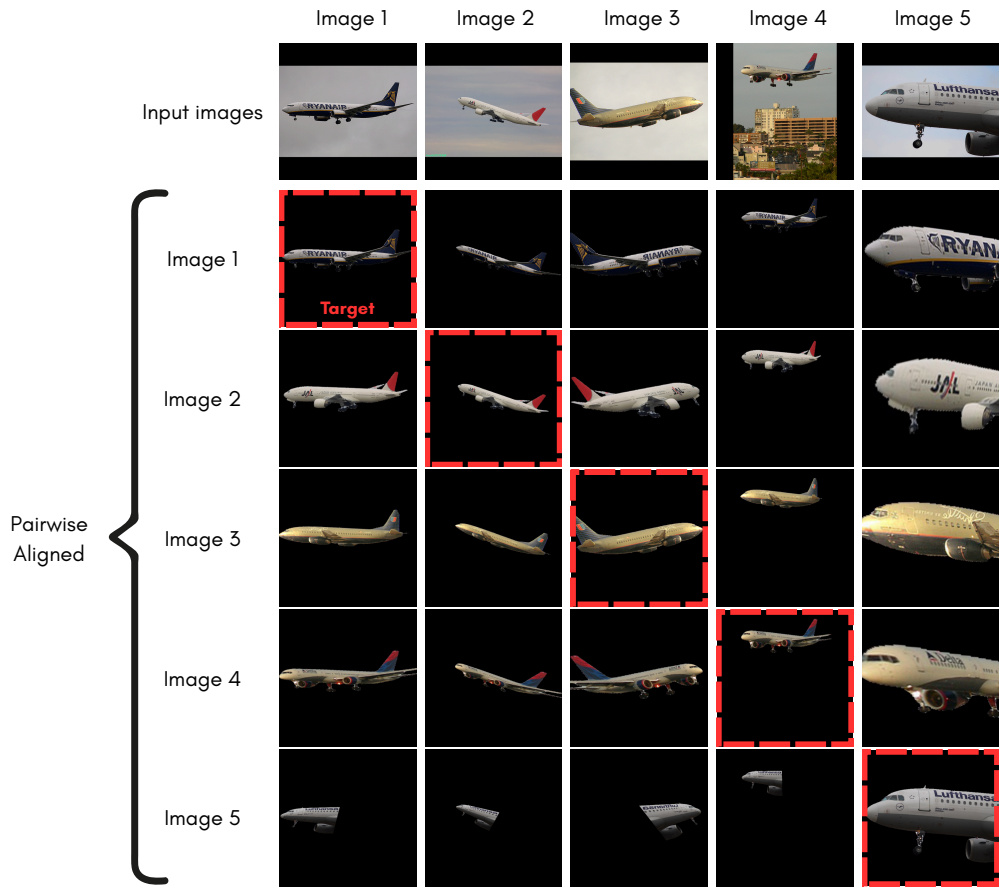

Figure 8: **Pairwise Alignment Grid – "Aeroplane" Class.** Top row: five input images from the "aeroplane" category. Below: a $5 \times 5$ alignment matrix, where each entry in column $j$ displays the corresponding source image from row $i$, warped to align with target image $j$ using the inverse-compositional transformation $T^{\theta_i} \circ T^{-\theta_j}$. Each row thus visualizes the same source image aligned to five different targets. Diagonal entries show the self-warped images (i.e., identity transformation), but with the canonical background rather than the original, and are marked with a **red** dashed-line frame. This layout highlights FastJAM's ability to achieve coherent, semantically meaningful alignments across all image pairs.

4

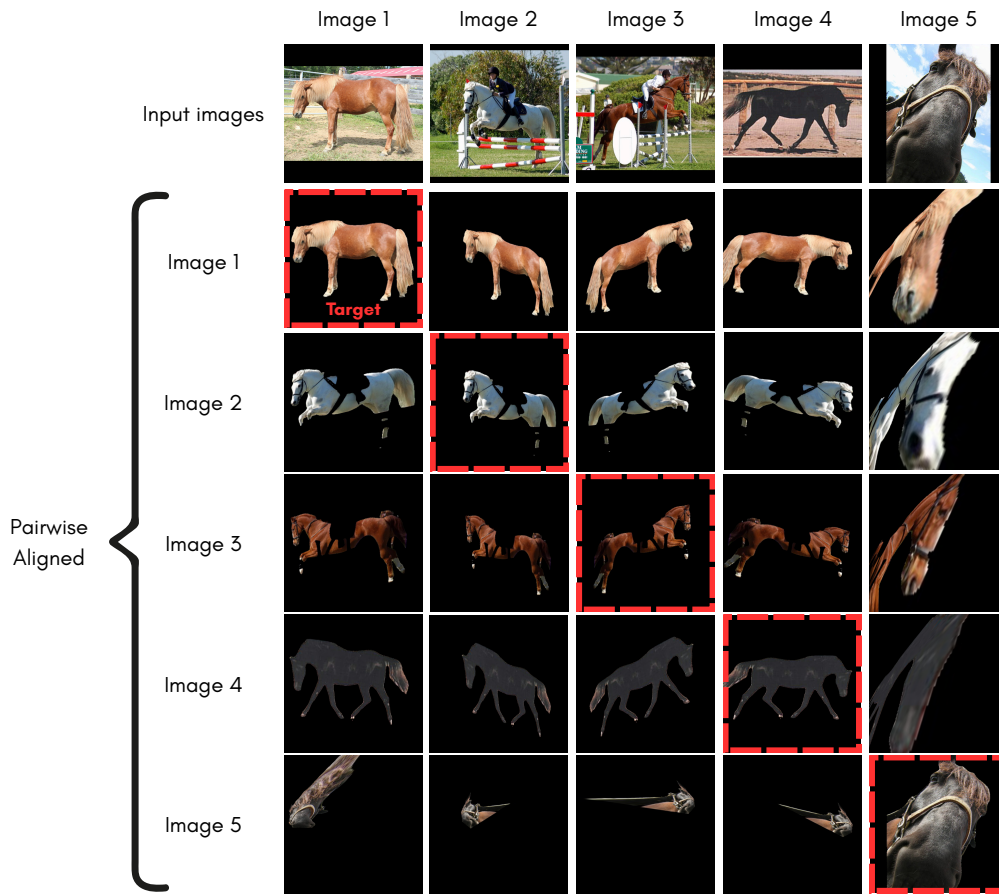

Figure 9: **Pairwise Alignment Grid – "Horse" Class.** Top row: five input images from the "horse" category. Below: a $5 \times 5$ grid showing the full pairwise alignment structure. Each image in row $i$, column $j$ corresponds to the source image $i$ aligned to the target image $j$ using $T^{\theta_i} \circ T^{-\theta_j}$. The diagonal entries depict self-warping (identity), rendered with the canonical background rather than the original, and are marked with a **red** dashed-line frame. This visualization reveals consistent alignment behavior across the set, illustrating how FastJAM handles pose and appearance variation within a semantic class.

5

## C Explaining the Colormap Visualization in More Detail

To provide an intuitive understanding of how FastJAM aligns images to a shared canonical space, we visualize the warped outputs using a predefined RGB colormap. As shown in Figure 10, each input image is first aligned to the canonical frame and blended with the colormap. The result is then inverse-warped back to the original image space, allowing us to visualize how semantic regions are mapped consistently across instances. This process highlights FastJAM's ability to establish meaningful correspondences without relying on dense features or explicit templates.

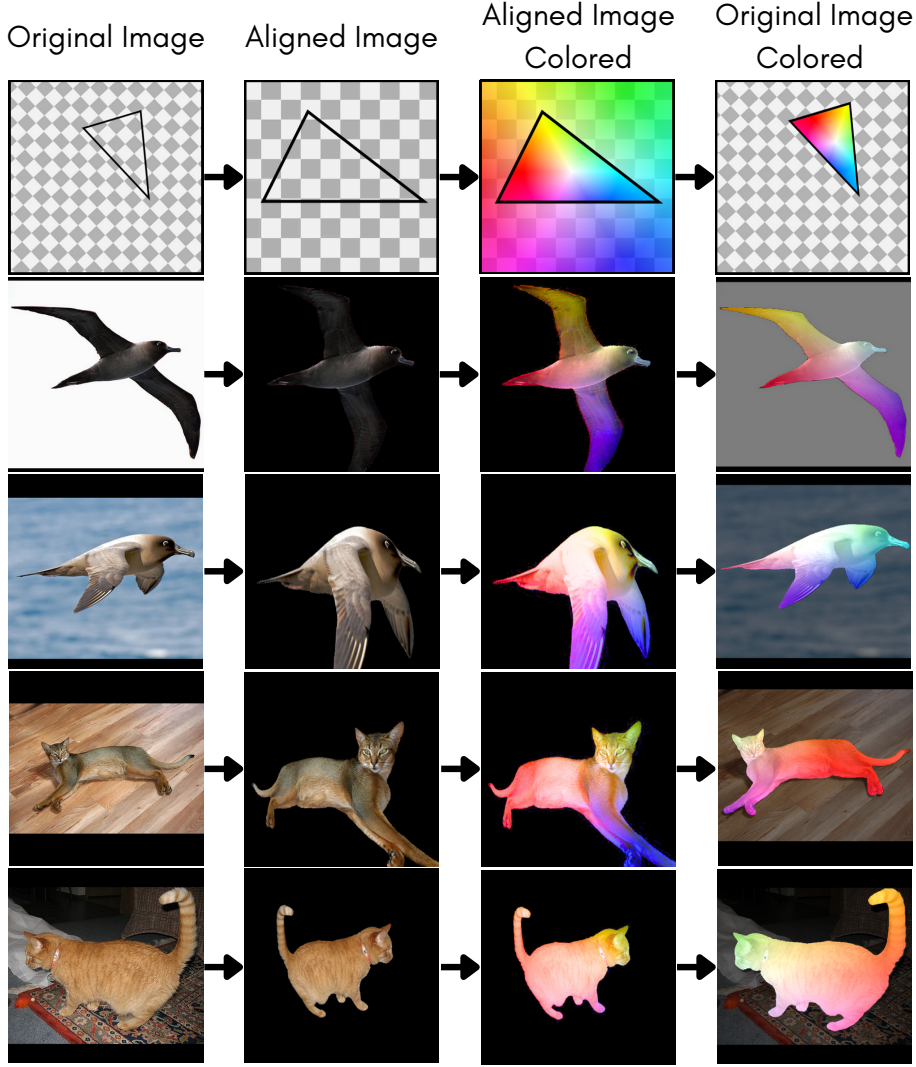

Figure 10: **Illustrated Canonical Space Visualization.** This figure provides an intuitive explanation of how FastJAM visualizes the canonical space $\mathcal{C}$ using a predefined RGB colormap $I_{\text{colormap}}$. For each input image $I_i$ (left column), we first apply the estimated homography to obtain its aligned version in canonical space: $I_i \circ T^{\theta_i}$. We then blend this aligned image with the colormap via averaging: $\frac{1}{2}(I_{\text{colormap}} + (I_i \circ T^{\theta_i}))$. Finally, we apply the inverse warp to visualize the blended canonical signal in the original image frame: $\left[\frac{1}{2}(I_{\text{colormap}} + (I_i \circ T^{\theta_i}))\right] \circ T^{-\theta_i}$. This provides a visual explanation of how semantically similar regions across instances are mapped to consistent spatial locations.

6

## D Runtime Analysis

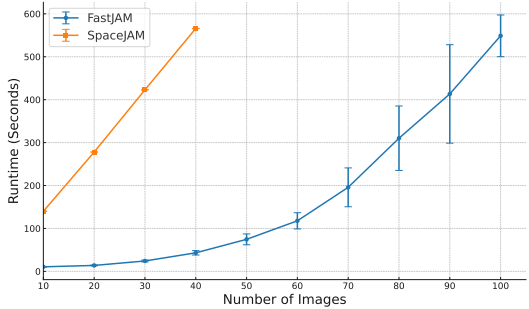

Figure 11: Runtime analysis between SpaceJAM and FastJAM over an increasing number of images. SpaceJAM PCA processing step runs out of RAM after 40 images.

As illustrated in Figure 11, FastJAM remains efficient across all tested sizes, completing alignment in under 450 seconds even for 100 images. In contrast, the current implementation of SpaceJAM runs out of RAM after $N = 40$ images due to the PCA preprocessing step.

## E Model Configuration and Training Setup

We train the model for 600 epochs using the Adam optimizer with an initial learning rate of $5 \times 10^{-3}$, multiplied by 0.5 after 200 epochs without improvement. The loss function is based on the Geman-McClure formulation with a robustness parameter $\sigma = 0.25$, and no weight decay is applied. The feature extractor within the GNN uses 5 layers of hidden size 128, followed by a linear projection to an 8-dimensional homography parameter vector. A full summary of the model architecture and its 133,256 trainable parameters is provided in Table 5. To encourage geometric stability, the final projection layer is initialized to approximate the identity transformation, and transformations are parameterized using Lie algebra to ensure invertibility. Alignment is applied iteratively using a single pass of the inverse compositional (IC) spatial transformer network. During optimization, horizontal flips are checked every 100 epochs.

Table 5: GraphSAGE GNN Model Summary.

| Layer (type:name) | Output Shape | Param |
|---|---|---|
| GraphSAGE GNN | – | – |
| convs.0.lin_l.weight | [128, 2] | 256 |
| convs.0.lin_l.bias | [128] | 128 |
| convs.0.lin_r.weight | [128, 2] | 256 |
| convs.1.lin_l.weight | [128, 128] | 16,384 |
| convs.1.lin_l.bias | [128] | 128 |
| convs.1.lin_r.weight | [128, 128] | 16,384 |
| convs.2.lin_l.weight | [128, 128] | 16,384 |
| convs.2.lin_l.bias | [128] | 128 |
| convs.2.lin_r.weight | [128, 128] | 16,384 |
| convs.3.lin_l.weight | [128, 128] | 16,384 |
| convs.3.lin_l.bias | [128] | 128 |
| convs.3.lin_r.weight | [128, 128] | 16,384 |
| convs.4.lin_l.weight | [128, 128] | 16,384 |
| convs.4.lin_l.bias | [128] | 128 |
| convs.4.lin_r.weight | [128, 128] | 16,384 |
| fc.weight | [8, 128] | 1,024 |
| fc.bias | [8] | 8 |
| **Total** | – | **133,256** |

For correspondence estimation, we employ the RoMa matcher at a fixed image resolution of $560 \times 560$ for both coarse and upsampled stages. A maximum of 10 keypoints is retained per image, filtered by

non-maximum suppression (NMS) using a radius of 0.054 in normalized coordinates—corresponding to a $30 \times 30$ pixel window in the original image space. This ensures spatial coverage while avoiding redundant detections.

Table 6: Summary of Notation

| Symbol | Description |
| --- | --- |
| $\mathcal{I} = (I_i)_{i=1}^N$ | Set of input images |
| $M_i$ | Number of keypoints in image $I_i$ |
| $X_i = \{x_i^{(1)}, \ldots, x_i^{(M_i)}\}$ | Keypoints in image $I_i$, $x_i^{(m)} \in [-1,1]^2$ |
| $\mathcal{M}_{ij}$ | Set of matched keypoints between $I_i$ and $I_j$ |
| $\mathcal{T}$ | Family of parametric transformations (e.g., homographies) |
| $T^{\boldsymbol{\theta}_i} \in \mathcal{T}$ | Transformation for image $I_i$, parameterized by $\boldsymbol{\theta}_i$ |
| $\mathcal{G} = (\mathcal{V}, \mathcal{E})$ | Keypoint graph with intra- and inter-image edges |
| $f(\mathcal{G})$ | GNN-based function that predicts $\{\boldsymbol{\theta}_i\}_{i=1}^N$ from a graph |
| $I_i \circ T^{\boldsymbol{\theta}_i}$ | Image $I_i$ warped by transformation $T^{\boldsymbol{\theta}_i}$ |

## F External Tools and Frameworks

We gratefully acknowledge the use of several open-source libraries and resources in this project. Our GNN implementation is based on torch_geometric, primarily using the GraphSAGE [4] architecture, along with other variants for comparison. We used Weights & Biases for experiment tracking and visualization. For keypoint matching, we built upon the official implementation of RoMa [5], and we also incorporated components from the LoFTR framework [6]. Object-centric masks were obtained using Grounded-SAM, which combines Grounding DINO [7, 8] and the Segment Anything Model (SAM) [9]. We thank the authors of all these works for making their code and models publicly available.

## G Lie Group Parameterization

A homography has 8 degrees of freedom and corresponds to an equivalence class of invertible matrices, where a representative with unit determinant can be used. Now consider

$$\mathfrak{sl}(3) = \left\{ \boldsymbol{\Theta} = \begin{bmatrix} \theta_1 & \theta_2 & \theta_3 \\ \theta_4 & \theta_5 & \theta_6 \\ \theta_7 & \theta_8 & -(\theta_1+\theta_5) \end{bmatrix} \right\} \quad \text{and} \quad \mathrm{SL}(3) = \left\{ \boldsymbol{H} \in \mathbb{R}^{3\times 3} : \det \boldsymbol{H} = 1 \right\}. \qquad (1)$$

The space $\mathfrak{sl}(3)$ is the Lie algebra of trace-zero matrices. The matrix exponential maps $\mathfrak{sl}(3)$ into $\mathrm{SL}(3)$, yielding a smooth eight-parameter representation of homographies. Our network predicts $(\boldsymbol{\theta}_i)_{i=1}^N$, where each $\boldsymbol{\theta}_i \in \mathbb{R}^8$ is mapped to $\boldsymbol{\Theta}_i \in \mathfrak{sl}(3)$ as shown above, and the homography is obtained via $T^{\boldsymbol{\theta}_i} = \boldsymbol{H}_i = \exp(\boldsymbol{\Theta}_i)$. This construction guarantees $\det T^{\boldsymbol{\theta}_i} = 1$. In particular, $\boldsymbol{\theta}_i = \boldsymbol{0}_{8\times 1}$ yields the identity matrix, and $T^{-\boldsymbol{\theta}_i}$ is the inverse of $T^{\boldsymbol{\theta}_i}$.



\end{document}